\documentclass{article}
\usepackage{ijcai25}

\usepackage{times}
\usepackage{soul}
\usepackage{url}
\usepackage[hidelinks]{hyperref}
\usepackage[utf8]{inputenc}
\usepackage[small]{caption}
\usepackage{graphicx}
\usepackage{svg}
\usepackage{amsmath}
\usepackage{amsthm}
\usepackage{booktabs}
\usepackage{algorithm}
\usepackage{algorithmic}
\usepackage[switch]{lineno}
\usepackage{amsfonts,amssymb}
\usepackage{colortbl}
\usepackage{marvosym}
\usepackage{subcaption}
\usepackage{multirow}
\newcommand{\xmark}{\ding{55}} 
\usepackage{pifont}   
\usepackage{makecell}
\usepackage{xcolor} 
\renewcommand{\arraystretch}{1.5} 

\nolinenumbers

\title{Concept-Aware Latent and Explicit Knowledge Integration for Enhanced Cognitive Diagnosis}

\author{
     Yawen Chen$^1$, Jiande Sun$^1$, Jing Li$^2$, Huaxiang Zhang$^1$
     \affiliations
$^1$School of Information Science and Engineering, Shandong Normal University\\
$^2$School of Journalism and Communication, Shandong Normal University\\
\emails
chenyawen111@163.com,
\{jiandesun, lijingjdsun\}@hotmail.com,
huaxzhang@163.com
}

\begin{document}

\maketitle

\begin{abstract}
     Cognitive diagnosis can infer the students’ mastery of specific knowledge concepts based on historical response logs, which makes it a fundamental upstream component of the intelligent education. However, the existing cognitive diagnostic models (CDMs) represent students' proficiency via a unidimensional perspective, which can’t assess the students’ mastery on each knowledge concept comprehensively. Moreover, the Q-matrix binarizes the relationship between exercises and knowledge concepts, and it can’t represent the latent relationship between exercises and knowledge concepts. Especially, when the granularity of knowledge attributes refines increasingly, the Q-matrix becomes incomplete correspondingly and the sparse binary representation (0/1) fails to capture the intricate relationships among knowledge concepts. To address these issues, we propose a Concept-aware Latent and Explicit Knowledge Integration model for cognitive diagnosis (CLEKI-CD). Specifically, a multidimensional vector is constructed according to the students' mastery and exercise difficulty for each knowledge concept from multiple perspectives, which enhances the representation capabilities of the model. Moreover, a latent Q-matrix is generated by our proposed attention-based knowledge aggregation method, and it can uncover the coverage degree of exercises over latent knowledge. The latent Q-matrix can supplement the sparse explicit Q-matrix with the inherent relationships among knowledge concepts, and mitigate the knowledge coverage problem. Furthermore, we employ a combined cognitive diagnosis layer to integrate both latent and explicit knowledge, further enhancing cognitive diagnosis performance. Extensive experiments on real-world datasets demonstrate that CLEKI-CD outperforms the state-of-the-art models across multiple evaluation metrics. The proposed CLEKI-CD is promising in practical applications in the field of intelligent education, as it exhibits good interpretability with diagnostic results.
\end{abstract}

\section{Introduction}

As a fundamental component of Intelligent Tutoring Systems (ITS) \cite{huang2020learning}, cognitive diagnosis evaluates students' mastery of specific knowledge concepts by analyzing their response records and is attracting increasing attention \cite{bu2022cognitive,lu2024design,liu2021towards,tong2021item}. A typical cognitive diagnostic system consists of three components: students, exercises, and knowledge concepts. Early classical cognitive diagnostic models, such as Item Response Theory (IRT) \cite{yen2006item} and Multidimensional Item Response Theory (MIRT) \cite{ackerman2014multidimensional}, offer a set of interpretable parameters that characterize both student abilities and exercise complexities. Subsequently, Deterministic-Input-Noisy-and-gate model (DINA) \cite{de2009dina} introduces the Q-matrix, which directly maps the representations of students and exercises to the corresponding knowledge concepts, allowing for the modeling of interactions between them. However, these traditional methods rely on manually-designed interaction functions, which are mathematical forms based on empirical and theoretical presuppositions, and are unable to adequately capture the complex nonlinear relationships between student abilities and exercise characteristics.

In recent years, numerous studies have been conducted to apply deep learning to the study of cognitive diagnosis \cite{cheng2019dirt,gao2023leveraging,huang2021group,ma2024dgcd}. NeuralCD \cite{wang2020neural} first integrated neural networks to learn complex high-order interactions among students, exercises, and knowledge concepts, making it a pioneering work. Since then, serval studies have tried to integrate graph neural network into CDMs to measure students' knowledge structures due to its advantages in capturing contextual information between nodes \cite{yang2016hierarchical,schlichtkrull2017modeling,wang2023self}. However, existing CDMs typically represent the students' mastery of a knowledge concept from a unidimensional perspective, i.e., in the form of scalar values. Considering that students' mastery of a certain concept is multi-perspective, encompassing aspects such as understanding, application, and extension, a single scalar value cannot fully capture the semantic information or establish interactions within hierarchical relationships. Therefore, using multidimensional vectors to represent a student's mastery of a knowledge concept is more intuitive and rational. Moreover, the Q-matrix represents the inclusion of knowledge concepts in exercises through a binary representation, improving the accuracy of inferred mastery and enhancing the interpretability of the CDMs. Nevertheless, as the granularity of knowledge attributes refines increasingly, the Q-matrix becomes progressively sparse and incomplete \cite{wang2022neuralcd}, and manually supplementing the Q-matrix is a laborious and expensive task. Additionally, the refining knowledge concepts and the limitations of students' time and interests, leading to insufficient knowledge concept coverage \cite{ma2022knowledge}. Therefore, leveraging the inherent relationships among knowledge concepts to uncover latent information and supplement the sparse explicit Q-matrix is essential for achieving a more comprehensive and reliable cognitive diagnosis.

To address the above issues, we propose a novel concept-aware latent and explicit knowledge integration model for cognitive diagnosis, termed CLEKI-CD, which achieves comprehensive cognitive diagnosis with interpretability. The main contributions of this work are listed as follows:
\begin{itemize}
\item We introduce a concept-aware multidimensional vector approach to represent students' mastery and exercise difficulty for each knowledge concept from multiple perspectives, which enhances representation capability by capturing richer semantic information on student characteristics and exercise attributes.
 \item We design an attention-based knowledge aggregation method to generate a latent Q-matrix representing implicit relationships between exercises and concepts to supplement the original sparse Q-matrix and alleviates the knowledge coverage problem. Building on this, we develop a combined cognitive diagnosis layer to integrate both types of knowledge, further enhancing cognitive diagnosis performance.
 \item Extensive experiments on real-world datasets demonstrate that CLEKI-CD outperforms current state-of-the-art models in terms of multiple metrics. Furthermore, case studies validate the interpretability and diagnostic efficacy of the model, showcasing its practical application in educational scenarios.
\end{itemize}

\section{Preliminaries}
\subsection{Task Overview}
In the domain of cognitive diagnosis, suppose a learning system consists of $N$ students, $M$ exercises  and $K$ knowledge concepts, where $\mathcal{S}=\{s_1, s_2, \ldots, s_N\}$, $\mathcal{E}=\{e_1, e_2, \ldots, e_M\}$, $\mathcal{K}=\{k_1, k_2, \ldots, k_K\}$, respectively. Each student selects exercises based on learning needs, which generates response logs in the form of a triplet $\mathcal{R} = \{(s, e, r) \mid s \in \mathcal{S}, e \in \mathcal{E}, r \in \{0, 1\}\}$, where $r$ represents the score indicating whether the response is correct $(1)$ or incorrect $(0)$. Additionally, the relationship between exercises and knowledge concepts is represented by the matrix $\mathbf{Q} = \{Q_{ij}\}_{M \times K}$, where each element $Q_{ij} = 1$ indicates a direct relevance of exercise \(e_i\) to concept \(k_j\), and $Q_{ij} = 0$ indicates no relevance. The Q-matrix is typically manually labeled by human experts in practice. Given the logs of students' responses $\mathcal{R}$ and the Q-matrix, the goal of our cognitive diagnostic task is to estimate student proficiency in each knowledge concept by predicting student performance on the new exercises. Like the exiting works \cite{wang2024survey}, we posit that when the calculated mastery level surpasses the corresponding difficulty of the knowledge concept, the student is more likely to answer the exercise correctly. Mathematically, we formulate the output of the framework as:

\begin{equation}
     y = \phi_n(\ldots \phi_1(h^{s}, h^{diff}, h^{other}, \Theta))
\end{equation}
where $\phi_i$ denotes the $i$-th ($i \in (0-n)$) MLP layer, $h^{s}$, $h^{diff}$, and $h^{other}$ represent the student's proficiency, the exercise difficulty in knowledge concepts, and other factors, respectively, and $\Theta$ denotes the model parameters.

\subsection{Concept Dependency Map}

\begin{figure}[tbp]
    \centering	\includegraphics[width=0.9\linewidth,height=0.5\textheight,keepaspectratio=true]{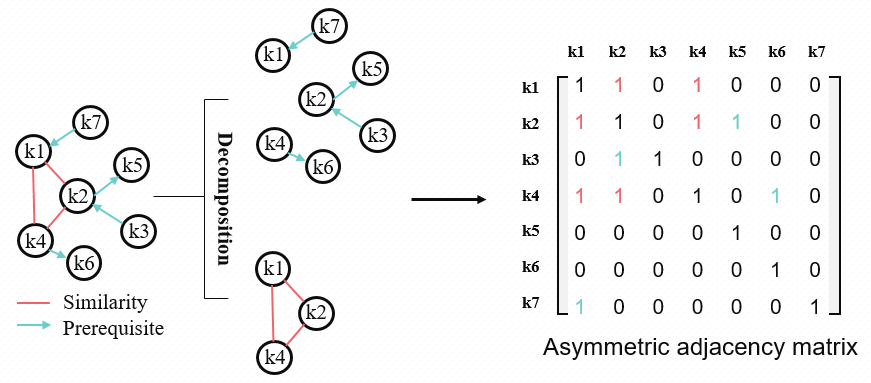}
    \caption{Decomposition of the concept dependency map and construction of the asymmetric adjacency matrix}
    \label{fig_2}
\end{figure}

 Educational theories indicate that various educational dependencies exist among knowledge concepts and the most typical are the prerequisite and similarity relations \cite{leighton2004attribute}. We design the concept dependency map based on expert-labeled educational dependencies, with a simple example in Figure \ref{fig_2}, where prerequisite relationships are directed and similarity relationships are undirected. As previously mentioned, if a student has mastered knowledge concept $k_2$, it is likely that he or she has also mastered the similar concepts $k_1$ and $k_4$, together with its prerequisite concept $k_3$. In this work, as shown in Figure \ref{fig_2}, we decompose the concept dependency map and construct an asymmetric adjacency matrix that distinguishes between two types of relationships: similarity (indicated by red undirected edges) and prerequisite (indicated by blue directed edges) dependencies. This matrix ensures that each node aggregates information solely from related nodes, and maintain an information flow consistent with the logical hierarchy of concepts.

\section{Method}
\begin{figure*}[t]
    \centering
	\includegraphics[width=0.8\linewidth, height=.6\textheight,keepaspectratio=true]{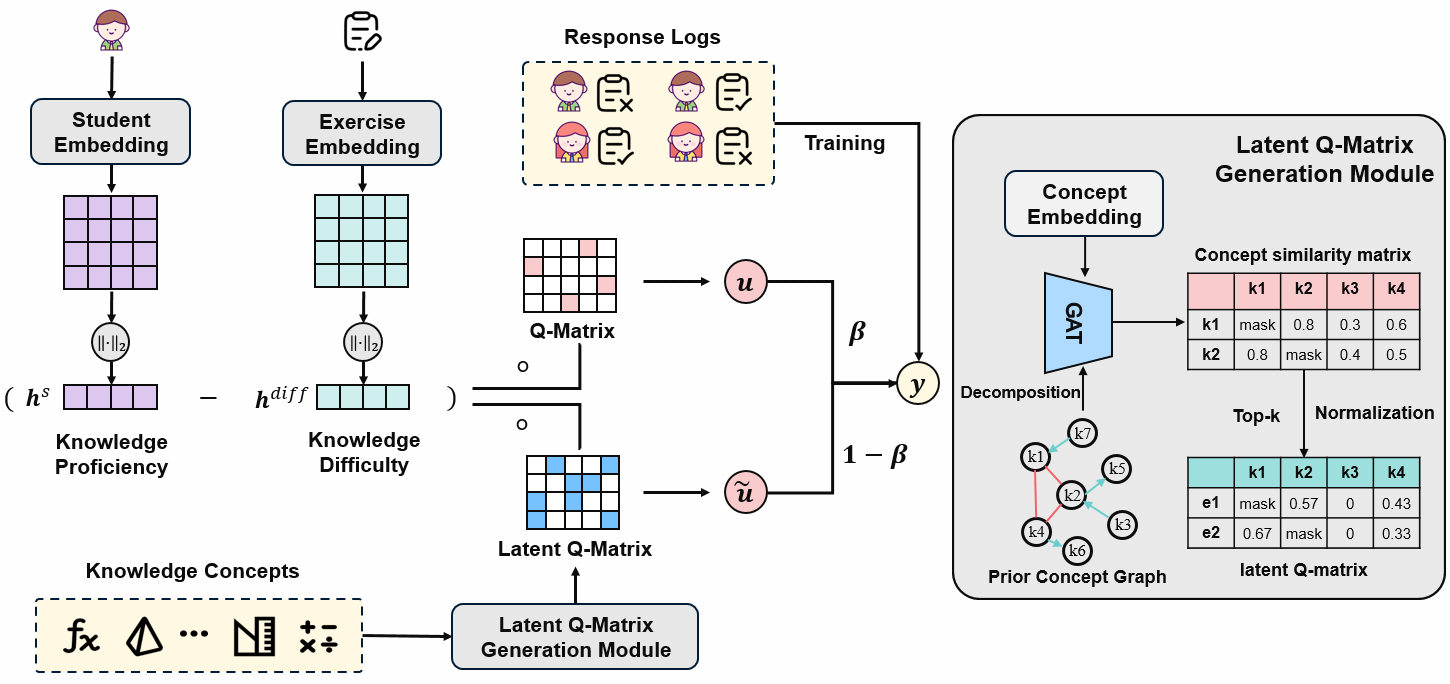}
    \caption{The overall structure of proposed CLEKI-CD. The Embedding Modules capture richer and more expressive features of student proficiency and exercise difficulty through vectorized representations. The Attention-Based Knowledge Aggregation Method generates a latent Q-matrix that complements the expert-defined explicit Q-matrix. The Combined Diagnostic Layer integrates both Q-matrices, enabling the model to leverage diverse cognitive concepts for a more comprehensive diagnosis.}
    \label{fig_3}
\end{figure*}

Our framework for cognitive diagnosis consists of multiple modules, as illustrated in Figure \ref{fig_3}. The following will introduce the details of each module.

\subsection{Concept-Aware Embedding Module}
Traditional methods model student proficiency and exercise difficulty from a unidimensional perspective, where each scalar value represents the level of mastery and difficulty of a specific concept. While this approach is intuitive, it is limited in representing student characteristics and exercise features. To address this limitation, we introduce a concept-aware embedding module by novelly extending the scalar representation into a multidimensional form, which helps capture richer semantic information and establishes interactions within hierarchical relationships. The details are as follows. 

First, we initialize the knowledge concept $k_c$ with its embedding representation vector $\mathbf{x}^k_c \in \mathbb{R}^D$. Accordingly, the embedding vectors for student $s_i$ and exercise $e_j$ are initialized as a concept-aware form as $\mathbf{x}^s_i \in \mathbb{R}^{K \times D}$ and $\mathbf{x}^e_j \in \mathbb{R}^{K \times D}$, where $K$ denotes the total number of knowledge concepts, and $D$ represents the dimensionality of the embedding space. Similar to existing neural network-based CDMs \cite{wang2020neural}, we use $\mathbf{h}^s$ and $\mathbf{h}^\text{diff}$ to characterize the student's proficiency on knowledge concepts and exercise difficulty, respectively, which can be obtained by:
\begin{equation}
\mathbf{h}^s_i = \sigma(\mathbf{x}_i^s \mathbf{W}_1  + \mathbf{b}_1)
\end{equation}
\begin{equation}
\mathbf{h}^{diff}_j = \sigma(\mathbf{x}_j^e \mathbf{W}_2  + \mathbf{b}_2)
\end{equation}
where $ \mathbf{W}_1 $ and $ \mathbf{W}_2 $ are the trainable weight matrices. To ensure the monotonicity assumption \cite{song2023deep} \cite{li2022cognitive} \cite{liu2024inductive}, which aligns with the real cognitive learning process, the elements of $ \mathbf{W}_1 $ and $ \mathbf{W}_2 $ are restricted to be positive. $ \sigma $ is the activation function, and here we use the sigmoid function.

\subsection{Latent Q-Matrix Generation via Attention-Based Knowledge Aggregation}
As the granularity of knowledge attributes is refined, the Q-matrix becomes increasingly sparse, which makes the diagnosis of more fine-grained knowledge concepts unreliable. This work assume that the Q-matrix provided by experts is always correct but may be incomplete \cite{liu2023qccdm}. We define the concepts labeled by experts in the original Q-matrix as explicit, while those not labeled are latent. We then explore the associations between two types of attributes to generate the latent Q-matrix, which complements the sparse original Q-matrix and mitigates the knowledge coverage problem. The specific process is as follows.

\noindent \textbf{Knowledge concept aggregation.} To generate the latent Q-matrix, an attention-based knowledge aggregation method is introduced. This method utilizes the topological relationships in the concept dependency graph and applies a Graph Attention Network (GAT) \cite{velivckovic2018graph} for knowledge concept aggregation. Unlike previous work that indiscriminately aggregates all knowledge embeddings, we distinguish between directed and undirected relationships among concepts and construct an asymmetric adjacency matrix, as shown in Figure \ref{fig_2}. For knowledge concept $k_i$, let $\mathbf{h}^k_i$ denote the corresponding representation embedding obtained from the first-order aggregation in the GAT, which can be computed as follows:
\begin{equation}
\mathbf{h}_i^k = \frac{1}{H} \sum_{h=1}^{H} \sum_{j \in \mathcal{N}(i)} \alpha_{ij}^h \mathbf{W}^h \mathbf{x}_j^k
\end{equation}
where $\mathcal{N}(i)$ represents the set of neighbor nodes of knowledge concept $k_i$, the representation $\mathbf{h}^k_i$ is derived by aggregating information from these neighbors. $\alpha_{ij}^h$ captures the dependency strength between knowledge concepts $\mathbf{x}_i^k$ and $\mathbf{x}_j^k$, thus preserving the dependency relationships within the concept map.
The dependency strength $\alpha_{ij}^h$ is defined as:

\begin{equation}
\small
\alpha_{ij}^h = \frac{\exp(\text{LeakyReLU}(\mathbf{a}^T [\mathbf{W}^h\mathbf{x}_i^h;\mathbf{W}^h\mathbf{x}_j^h]))}{\sum_{n \in \mathcal{N}(i)} \exp(\text{LeakyReLU}(\mathbf{a}^T [\mathbf{W}^h\mathbf{x}_i^h;\mathbf{W}^h\mathbf{x}_n^h]))}
\end{equation}
where $\mathbf{W}^h$ is a trainable weight matrix applied to node features and $\mathbf{a}$ is a learnable attention vector used to compute attention coefficients. The $[\cdot ; \cdot]$ represents concatenation operation.

\noindent \textbf{Latent Q-matrix generation.} Following the information aggregation via GAT, we derive the aggregated knowledge embedding matrix $\mathbf{H}^{k} = [\mathbf{h}_1^{k}, \mathbf{h}_2^{k}, \ldots, \mathbf{h}_K^{k}]^{T} \in \mathbb{R}^{K\times D}$. To further identify the latent knowledge concepts associated with the explicit concept attributes, we employ cosine similarity to measure the relationship between each explicit knowledge concept and all latent knowledge concepts linked to it, as defined by the concept dependency map. The similarity matrix calculation process is illustrated below:

\begin{equation}
    \mathbf{S} = \mathbf{M}  \left( \frac{\mathbf{H}^{k} \mathbf{H}^{kT}}{\|\mathbf{H}^{k}\|_2 \|\mathbf{H}^{kT}\|_2} \right)
\end{equation}
where $\mathbf{M} \in \mathbb{R}^{K \times K}$ is the mask matrix designed to exclude self-similarities and unrelated knowledge concepts, and it is defined as:

\begin{equation}
\small
    M_{ij} = 
\begin{cases} 
0, & \text{if knowledge } i \text{ and } j \text{ are not associated}, \\
1, & \text{if knowledge } i \text{ and } j \text{ are associated and } i \neq j.
\end{cases}
\end{equation}

To emphasize the most relevant similarities, we select the top-$P$ highest similarity values in each row of $\mathbf{S}$, apply a softmax operation to these values, and set the remaining entries to zero to maintain the matrix's original dimensions:

\begin{equation} 
\tilde{S}_{ij} = \mathbb{I}{{j \in \mathcal{P}_i}} \times \frac{\exp(S_{ij})}{\sum_{k \in \mathcal{P}_i} \exp(S_{ik})} \end{equation} 
where $\mathcal{P}_i$ is the set of indices corresponding to the top-$P$ highest similarity values in the $i$-th row of $\mathbf{S}$, and $\mathbb{I}{{j \in \mathcal{P}_i}}$ is the indicator function that equals $1$ if $j \in \mathcal{P}_i$, and $0$ otherwise. Using this refined similarity matrix, we define the latent matrix $\tilde{\mathbf{Q}}$ as:

\begin{equation}
    \tilde{\mathbf{Q}} = \mathbf{Q}\tilde{\mathbf{S}}
\end{equation}

Using the attention-based knowledge aggregation method, we construct a latent Q-matrix that supplements the original, alleviating the knowledge coverage issue.

\subsection{Combined Cognitive Diagnosis with Explicit and Latent Knowledge}
Existing CDMs typically depend solely on the Q-matrix to identify the knowledge concepts each exercise involves. However, due to potential biases in manual labeling and the inherent knowledge coverage limitations, it is essential to account for the influence of latent knowledge concepts as well. In this subsection, we propose a combined diagnosis layer that integrates both explicit and latent Q-matrices, enabling the model to leverage both types of knowledge for a more comprehensive and accurate diagnosis. 

\noindent \textbf{Diagnosis based on explicit Q-matrix.}
The explicit Q-matrix provides a structured mapping of the relations between exercises and directly associated concepts. This mapping allows us to evaluate a student’s performance on each exercise by focusing on the explicitly examined knowledge concepts. For the $j$-th exercise, the corresponding knowledge concept code $\mathbf{q}_j$ is obtained by $\mathbf{q}_j = \mathbf{e}_j \times \mathbf{Q}$, where $\mathbf{e}_j \in \mathbb{R}^{1 \times M}$ is the one-hot encoding of the $j$-th exercise. In the diagnosis layer, the interaction function for the explicit knowledge concepts in the explicit Q-matrix is defined as:
\begin{equation}
u_{ij} = \sigma(\frac{1}{c_j} \sum_{k=1}^{K} q_j^k \times (\|\mathbf{h}^s_i\|_2 - \|\mathbf{h}^{diff}_j\|_2))
\end{equation}
where $c_j$ is the number of explicit knowledge concepts included in the exercise, $q_j^k$ is the $k$-th knowledge concept in $\mathbf{q}_j$, and $||\cdot||_2$ denotes the $L_2$ norm operation.

\noindent \textbf{Diagnosis based on latent Q-matrix.} Similar to the diagnosis process based on the explicit Q-matrix, the process for computing the diagnosis results using the latent Q-matrix is formulated as follows:
\begin{equation}
\tilde{u}_{ij} = \sigma(\frac{1}{c_j} \sum_{k=1}^{K} \tilde{q}_j^k \times (\|\mathbf{h}^s_i\|_2 - \|\mathbf{h}^{diff}_j\|_2))
\end{equation}
where $\tilde{q}_j^k$ is the $k$-th latent knowledge concept in $\tilde{\mathbf{q}}_j$. For the $j$-th exercise, the corresponding latent knowledge concept $\tilde{\mathbf{q}}_j$ is obtained by $\tilde{\mathbf{q}}_j = \mathbf{e}_j \times \tilde{\mathbf{Q}}$.

\noindent \textbf{Combined Diagnosis Layer.} The final prediction of the model is a weighted sum of the predictions from both explicit and latent knowledge components, which can be expressed as:
\begin{equation}
    y_{ij} = \epsilon u_{ij} + (1 - \epsilon) \tilde{u}_{ij}
\end{equation}
where $\epsilon$ is the coefficient representing the contribution of predictions from both types of knowledge. This balancing factor allows the model to dynamically integrate information from both explicit and latent knowledge, enhancing the robustness and rationality of cognitive diagnosis.

\subsection{Model Optimization}

To ensure the model is effectively optimized, we adopt a loss function based on cross-entropy between the predicted output and the actual response. For a given student $s_i$, exercise $e_j$, and predicted response $y_{ij}$, the loss function is defined as follows:
\begin{equation}
\small
     \mathcal{L} = - \sum_{(s_i, e_j, r_{ij}) \in \mathcal{R}} \left( r_{ij} \log y_{ij} + (1 - r_{ij}) \log (1 - y_{ij}) \right)
\end{equation}
where $r_{ij}$ is the true response and $\mathcal{R}$ is the set of all student-exercise-response triples.

\section{Experiments}

\subsection{Experimental settings}
\noindent \textbf{Datasets.} We conduct experiments on two real-world datasets, i.e., ASSIST \cite{wang2020neural} and Junyi \cite{chang2015modeling}, which widely used for cognitive diagnostic tasks \cite{shao2024improving}. During preprocessing, we remove students with fewer than 15 logs in each dataset to ensure sufficient data for conducting effective diagnosis for each student. The detailed statistics are recorded in Table \ref{table1}. It is worth noting that both "Sparsity in student-concept interactions" and "Sparsity in student-exercise logs" indicate that most students have completed only a limited number of exercises, resulting in minimal interaction with many knowledge concepts, which leads to high sparsity of data.

\begin{table}[tb]
\centering
\footnotesize
\renewcommand{\arraystretch}{0.8}
\begin{tabular}{lcc}
\toprule
\textbf{Dataset}                           & \textbf{Junyi} & \textbf{Assist} \\ 
\midrule
\#Students                                    & 10,000         & 2,493           \\ 
\#Exercises                                   & 734            & 17,746          \\ 
\#Knowledge concepts                          & 734            & 123             \\ 
\#Response records                            & 408,057        & 267,415         \\ 
\#Response logs per student                   & 40.8           & 107.266         \\
\#Knowledge concepts per exercise             & 1              & 1.192           \\
\#Sparsity in student-exercise logs           & 94.44\%       & 99.39\%        \\
\bottomrule
\end{tabular}
\caption{The statistics of the datasets}
\label{table1}
\end{table}

\noindent \textbf{Evaluation Metrics.} Due to the lack of access to students' true mastery of knowledge concepts, following previous studies \cite{liu2018fuzzy,gao2021rcd,li2022hiercdf,wang2023self} , we assess models by predicting students' performance on the test set of response logs and use common evaluation metrics including Accuracy (ACC), Area Under the Curve (AUC) \cite{myerson2001area}, and Root Mean Square Error (RMSE) \cite{willmott2005advantages}.

\subsection{Experimental Results}

To provide a thorough assessment of the effectiveness of our model, we randomly split the datasets into training sets of 80\%, 70\%, and 60\%, and compare it against the baselines including IRT, MIRT, NeuralCD, KSCD \cite{ma2022knowledge}, RCD, KANCD \cite{wang2022neuralcd} and ORCDF \cite{qian2024orcdf}. Table \ref{table:performance} presents the experimental results with the best scores highlighted in bold. Our model achieves the best performance across all train-test splits on two datasets and traditional psychology-based cognitive diagnostic models like IRT and MIRT perform the weakest. NeuralCD and KANCD first simulate complex student-exercise interactions using neural networks and attempt to address the knowledge coverage problem but do not explore the implicit relationships between concepts. KSCD infers the mastery of unpracticed concepts by learning intrinsic relationships between concepts, but indiscriminately aggregates all knowledge embeddings. RCD and ORCDF model interactions through a student-practice-concept relationship graph but do not address the sparsity and incompleteness of the Q-matrix. CLEKI-CD further enhances representation capability by concept-aware multidimensional vector and generates latent knowledge concepts to address knowledge coverage. This approach provides a more comprehensive and interpretable framework.

\begin{table*}[tb]
    \centering
    \resizebox{\textwidth}{!}{
    \begin{tabular}{l|ccc|ccc|ccc|ccc|ccc|ccc}
        \toprule
          & \multicolumn{9}{c|}{\textbf{Assist Dataset}} & \multicolumn{9}{c}{\textbf{Junyi Dataset}} \\
        \midrule 
        \multirow{2}{*}{\textbf{Methods}} & \multicolumn{3}{c|}{\textbf{80\%/20\%}} & \multicolumn{3}{c|}{\textbf{70\%/30\%}} & \multicolumn{3}{c|}{\textbf{60\%/40\%}} & \multicolumn{3}{c|}{\textbf{80\%/20\%}} & \multicolumn{3}{c|}{\textbf{70\%/30\%}} & \multicolumn{3}{c}{\textbf{60\%/40\%}} \\
        & \textbf{ACC}   & \textbf{AUC}   & \textbf{RMSE}  & \textbf{ACC}   & \textbf{AUC}   & \textbf{RMSE}  & \textbf{ACC}   & \textbf{AUC}   & \textbf{RMSE} & \textbf{ACC}   & \textbf{AUC}   & \textbf{RMSE}  & \textbf{ACC}   & \textbf{AUC}   & \textbf{RMSE}  & \textbf{ACC}   & \textbf{AUC}   & \textbf{RMSE} \\
        \midrule
        IRT    & 68.72  & 69.77  & 45.37  & 67.19  & 72.93  & 47.75  & 66.21  & 71.89  & 48.51  & 71.82  & 74.76  & 45.89  & 70.31  & 73.52  & 46.95  & 70.16  & 74.63  & 44.31 \\
        MIRT   & 70.08  & 72.76  & 48.88  & 69.06  & 73.47  & 45.58  & 69.90  & 74.15  & 45.20  & 72.14  & 75.88  & 44.83  & 71.89  & 74.08  & 45.90  & 70.03  & 73.51  & 44.21 \\
        NeuralCD   & 71.83  & 74.30  & 44.33  & 71.05  & 73.49  & 44.13  & 68.92  & 72.45  & 47.12  & 74.17  & 77.16  & 42.83  & 73.60  & 76.21  & 44.64  & 71.15  & 75.15  & 45.67 \\
        KSCD & 71.05  & 73.51  & 45.32  & 70.71  & 72.76  & 45.62  & 69.52  & 73.01  & 46.10  & 73.87  & 76.66  & 43.62  & 72.40  & 74.14  & 43.88  & 71.54  & 75.38  & 48.49 \\
        RCD   & 72.00  & 76.40  & 43.26  & 72.09  & 75.57  & 43.33  & 70.10  & 75.02  & 43.25  & 75.86  & 78.76  & 41.88  & 73.49  & 76.87  & 44.16  & 72.08  & 74.19  & 42.44 \\
        KANCD  & 72.31  & 75.77  & 43.51  & 71.63  & 75.80  & 43.35  & 71.50  & 75.65  & 43.05  & 75.51  & 78.67  & 42.09  & 74.05  & 77.41  & 42.32  & 73.62  & 77.28  & 42.63 \\
        ORCDF  & 73.09  & 76.33  & 42.25  & 72.35  & 75.97  & 42.95  & 72.85  & 76.02  & 42.76  & 76.49  & 80.11  & 40.42  & 75.08  & 79.27  & 41.47  & 74.15  & 75.97  & 44.32 \\
        \textbf{Ours}   & \textbf{74.35}  & \textbf{77.89}  & \textbf{41.70}  & \textbf{73.80}  & \textbf{77.07}  & \textbf{42.10}  & \textbf{73.52} & \textbf{76.66} & \textbf{42.34} & \textbf{77.95}  & \textbf{82.03}  & \textbf{39.01}  & \textbf{76.50}  & \textbf{81.20}  & \textbf{40.05}  & \textbf{75.84}  & \textbf{78.62}  &  \textbf{41.90} \\
        \bottomrule
    \end{tabular}
    }

    \caption{Experimental results on student performance prediction. Left: Assist Dataset; Right: Junyi Dataset.}
    \label{table:performance}
\end{table*}

Table \ref{table:performance} shows that the model performs better on all metrics on the Junyi dataset. To further analyze, we create a new Junyi-part dataset, ensuring that the number of students and response logs are comparable to ASSIST. Table \ref{tab:data-difference} shows that the accuracy on the Junyi-part is lower than the original Junyi dataset due to the insufficient number of response logs, which prevents the model from accurately learning the students' answer patterns. On the other hand, although the number of students and responses are similar, the ACC achieved in ASSIST is obviously lower than that in Junyi-part. It is because that larger pool of exercises in ASSIST leads to sparser student-exercise interactions than Junyi-Part, which limits the ability to effectively learn the relationships between students and knowledge concepts. Therefore, educators should strengthen student-exercise interactions in teaching to ensure that students are exposed to a wider range of knowledge concepts.

\begin{table}[ht]
\centering
\resizebox{0.9\linewidth}{!}{
\begin{tabular}{c|ccccc}
\hline
  & \textbf{Student} & \textbf{Exercise} & \textbf{Response logs
} & \textbf{Sparsity} & \textbf{ACC} \\ \hline
Assist & 2493 & 17,746 & 267,415 & 99.39\% & 74.35 \\ 
Junyi  & 10,000 & 734 & 408,057 & 94.44\% & 77.95 \\ 
Junyi-part & 2493 & 734 & 216,562 & 88.17\% & 75.75 \\ \hline
\end{tabular}
}
\caption{Model performance and statistics across three datasets}
\label{tab:data-difference}
\end{table}

\subsection{Ablation Study}

We conduct ablation studies on the ASSIST dataset to evaluate the impact of each module of CLEKI-CD. The four variants of CLEKI-CD include CLEKI-CD w/o MRP (multidimensional representation of student proficiency and exercise difficulty on concepts), CLEKI-CD w/o AGM (the attention-based knowledge aggregation module), CLEKI-CD w/o CD-LK (the diagnostic layer with latent knowledge) and CLEKI-CD w/o CD-EK (the cognitive diagnosis with explicit knowledge). We assess the performance of variants with a consistent experimental configuration. The results in Table \ref{table-ablation} show that the ACC of all variants is below 74\%, which highlights the pivotal contribution of each module. Specifically, the MRP enhances the ability to capture student and exercise characteristics. The AGM dynamically weights neighboring concepts to capture prerequisite and similarity relations, contextualizing each concept within the broader knowledge structure for accurate diagnosis inferences. The CD-EK module evaluates student performance based on explicit knowledge from the original Q-matrix, providing a reliable foundation for accurately gauging mastery over well-defined, expert-labeled concepts. Finally, the CD-LK module supplements the sparse and incomplete Q-matrix by inferring latent knowledge relationships, enhancing predictive robustness in real-world settings with previously unmeasured concepts.

\begin{table}[tb]
\centering
\resizebox{0.9\linewidth}{!}{
\begin{tabular}{lccccccc}
\hline
\textbf{Variants} & MRP & AGM & CD-LK & CD-EK & ACC & AUC & RMSE \\ 
\hline
Variant1     & \xmark & \checkmark  &  \checkmark  &  \checkmark  & 73.47  & 77.21  & 42.49  \\
Variant2    &  \checkmark &  \xmark & \checkmark  & \checkmark  & 73.24  & 76.49  &	42.78  \\
Variant3   &  \checkmark & \checkmark  &  \xmark  & \checkmark  &  73.69 & 76.34  & 42.31  \\
Variant4   &  \checkmark &  \checkmark &  \checkmark  & \xmark   &  73.99 & 77.65  & 41.83  \\
CLEKI-CD   & \checkmark  & \checkmark  & \checkmark  & \checkmark  & \textbf{74.35}	 & \textbf{77.89}  & \textbf{41.70}  \\    
\hline 
\end{tabular}
}
\caption{Results of ablation experiment on Assist dataset}
\label{table-ablation}
\end{table}

\subsection{Case Study}
\begin{figure}[htbp]
    \centering
    \begin{subfigure}{0.49\linewidth}
		\centering
		\includegraphics[width=1.0\linewidth]{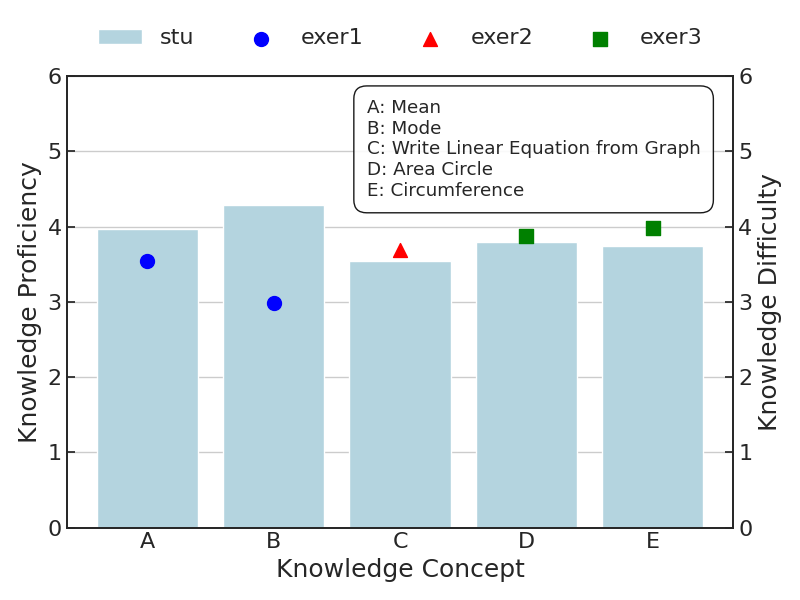}
		\caption{}
		\label{fig:acc1}
	\end{subfigure}
    \begin{subfigure}{0.49\linewidth}
		\centering
		\includegraphics[width=1.0\linewidth]{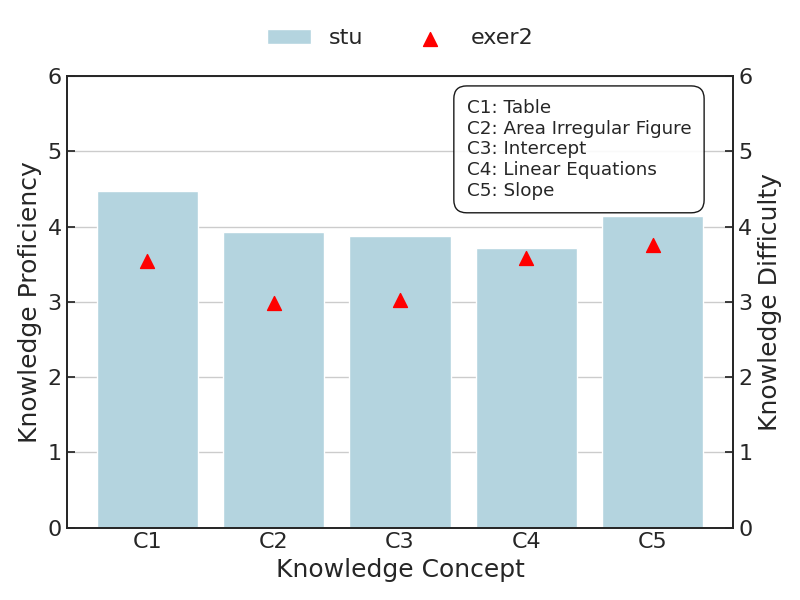}
		\caption{}
		\label{fig:acc2}
	\end{subfigure}

    \caption{Proficiency and difficulty on (a) knowledge concepts A-E, and (b) the top 5 latent concepts associated with concept C .}
    \label{fig_4}
\end{figure}

To demonstrate the interpretability of the model, we conduct a case study using three exercises from the ASSIST dataset. Table \ref{table4} presents the Q-matrix for these exercises along with the student’s actual responses. Figure \ref{fig_4} (a) illustrates the diagnostic results, where the bars depict the student proficiencies in each knowledge concept, and colored markers indicate the concept difficulty. For Exercise 1, the model predicts that the student's proficiency in two concepts exceeds their difficulty, aligns with the actual response (R). The same conclusion applies to Exercise 3, where the student’s proficiency in the two concepts is below their difficulty, and the actual response is W. However, the diagnosis of Concept C does not match the actual result for Exercise 2. In our proposed method, the diagnosis of Concept C is presented by the proficiency and difficulty of the five latent knowledge concepts most relevant to it, and Figure \ref{fig_4} (b) indicate that the proficiency of these concepts is higher than their difficulty. The diagnosis obtained by our proposed method can match to the actual response. This suggests that despite the explicit concept's proficiency being slightly below its difficulty, the higher proficiency of related latent concepts supports the correct diagnosis. These findings highlight the auxiliary role of latent knowledge concepts in enhancing the predictive outcomes of the model.

\begin{table}[ht]
\centering
\resizebox{0.65\linewidth}{!}{
\footnotesize
\begin{tabular}{ccccccc}
    \hline
     & A & B & C & D & E & \textbf{Response} \\ \hline
    Exercise 1 & 1 & 1 & 0 & 0 & 0 & \textcolor{green}{R} \\ \hline
    Exercise 2 & 0 & 0 & 1 & 0 & 0 & \textcolor{green}{R} \\ \hline
    Exercise 3 & 0 & 0 & 0 & 1 & 1 & \textcolor{red}{W} \\ \hline
\end{tabular}
}
\caption{Knowledge concepts associated with exercises and the true response for each student-exercise interaction}
\label{table4}
\end{table}

\subsection{Embedding Visualization}

\begin{figure}[tb]
    \centering
    \begin{subfigure}{0.49\linewidth}
		\centering
		\includegraphics[width=1.0\linewidth]{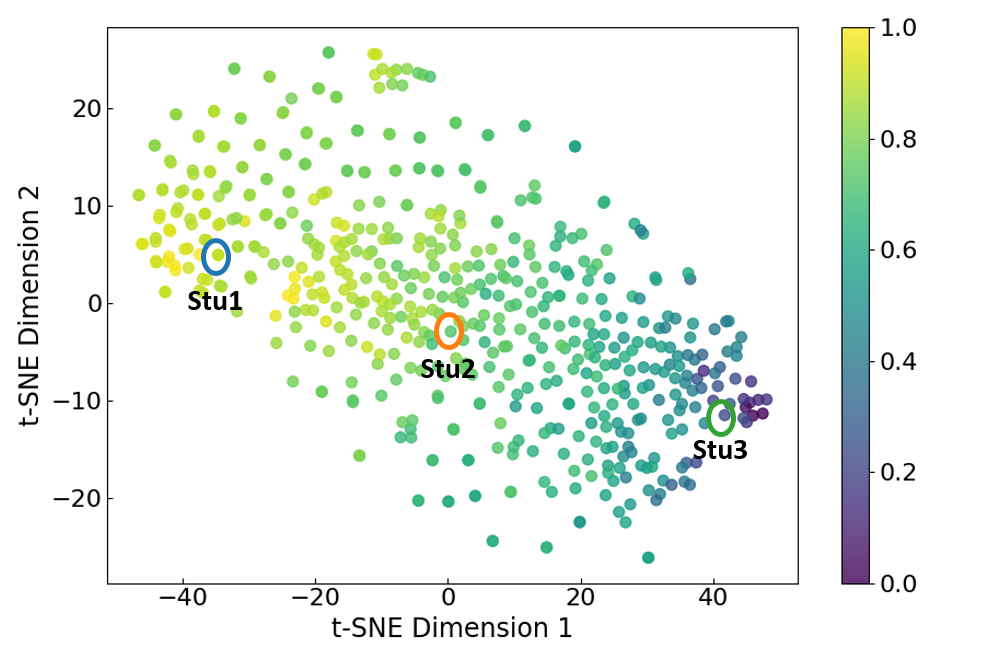}
		\caption{}
		\label{fig:stu_tsne}
	\end{subfigure}
    \begin{subfigure}{0.49\linewidth}
		\centering
		\includegraphics[width=1.0\linewidth]{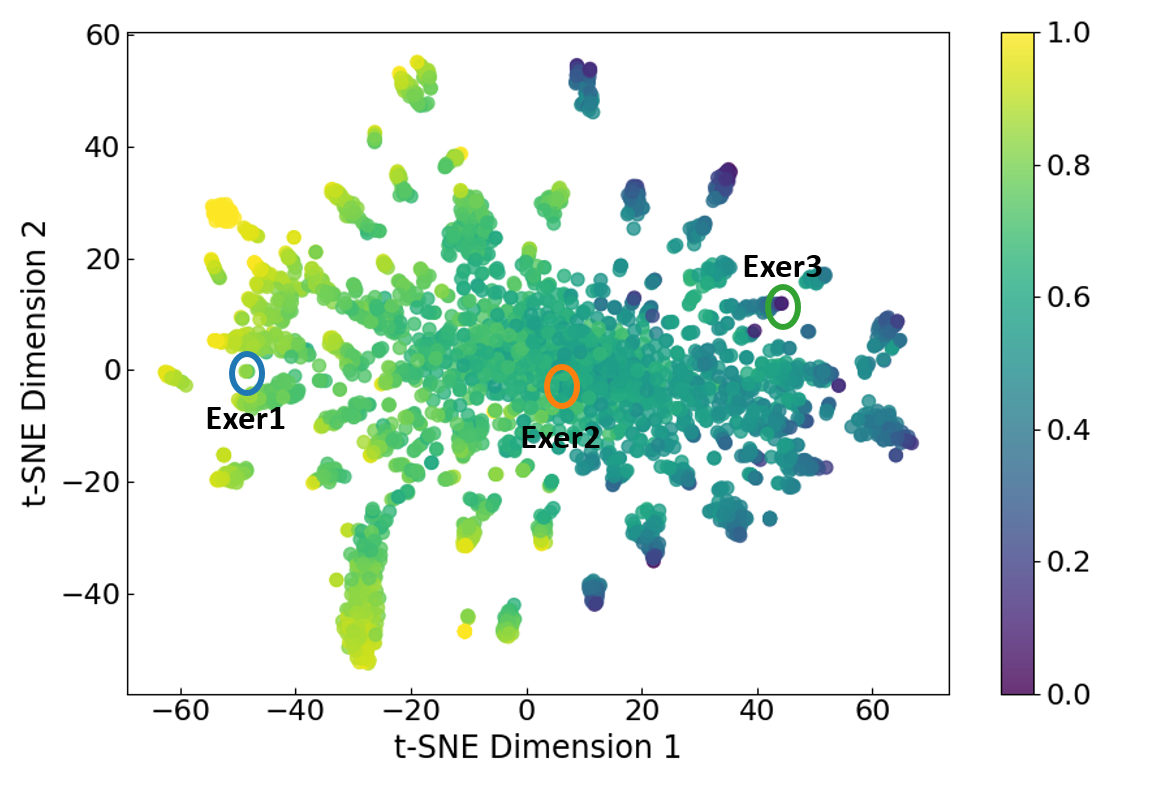}
		\caption{}
		\label{fig:exer_tsne}
	\end{subfigure}
    \\
    \begin{subfigure}{0.49\linewidth}
		\centering
		\includegraphics[width=1.0\linewidth]{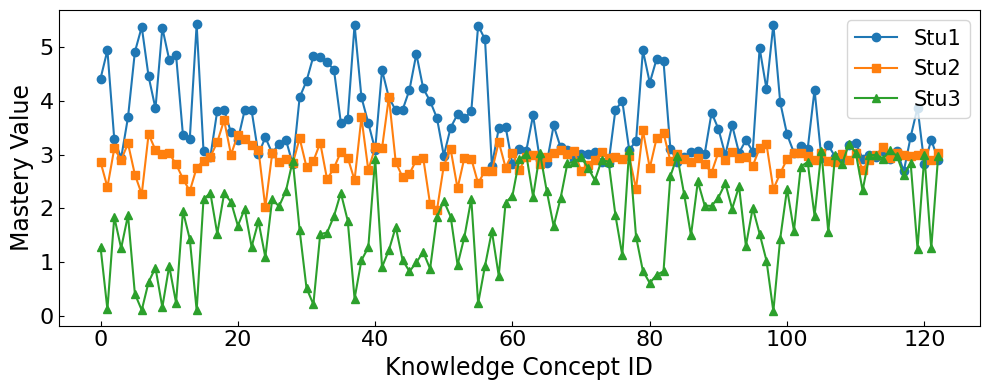}
		\caption{}
		\label{fig:stu-mas}
	\end{subfigure}
    \begin{subfigure}{0.49\linewidth}
		\centering
		\includegraphics[width=1.0\linewidth]{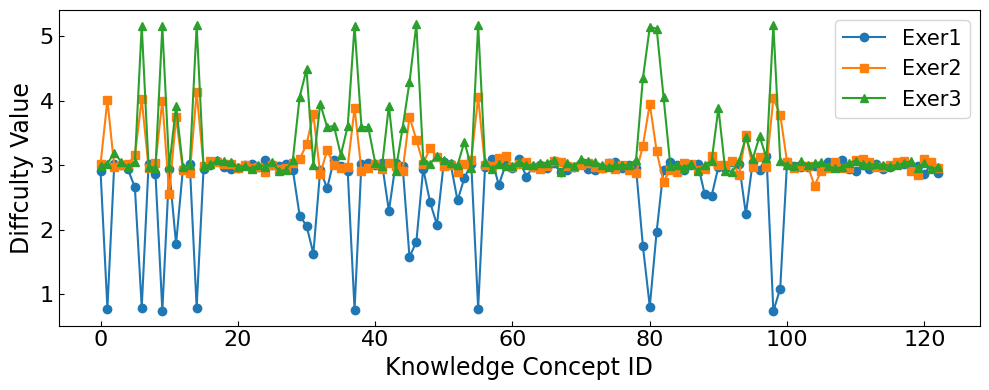}
		\caption{}
		\label{fig:exer_diff}
	\end{subfigure}

    \caption{The t-SNE visualization of (a) students' mastery and (b) exercise difficulty. Examples of diagnostic results on knowledge concepts of (c) students' proficiency and (d) exercise difficulty.}
    \label{fig_5}
\end{figure}

Our model employs vector embeddings to represent the mastery level and difficulty of each concept. We apply the t-SNE method to verify that the model effectively learn distinguishable representations, where the colors represent the normalized correct rates. To ensure the reliability of the visualizations, we focus on students with more than 100 responses and exercises with more than 20 responses, as shown in Figure \ref{fig_5} (a) and (b). 

Figure \ref{fig_5} (a) and (b) demonstrate the discernible clustering of point distributions rather than completely random scatter. High-level students (closer to yellow) exhibit a higher mastery level across most knowledge concepts, while low-level students (closer to blue) show the opposite, thus forming two clusters. Mid-level students display varying mastery across different concepts, resulting in a more dispersed distribution. Figure \ref{fig_5} (b) indicates that exercises of moderate difficulty are more concentrated in the embedding space, as they typically encompass knowledge concepts that align with the capabilities of the majority of students. In contrast, exercises with either low or high difficulty emphasize basic or complex concepts, respectively, causing their embeddings to be biased toward the ends of the space. Additionally, the results presented in Figure \ref{fig_5} (c) and (d) further confirm that the selected samples correspond to the characteristics of their respective areas. Overall, this model effectively learns distinguishable representations, thereby enhancing the interpretability of cognitive diagnostic results.

\subsection{Evaluation of Knowledge Coverage Robustness}

\begin{figure}[tb]
    \centering
	\includegraphics[width=.7\linewidth, height=.5\textheight,keepaspectratio=true]{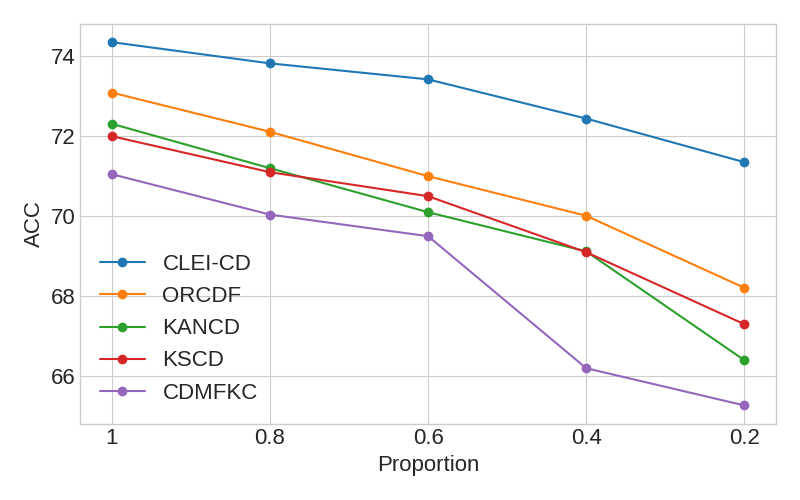}
    \caption{ACC of models under varying training data proportions.}
    \label{fig_few_shot}
\end{figure}

To evaluate the effectiveness of our model in alleviating the knowledge coverage problem, we conduct experiments under varying knowledge coverage rates, as shown in Figure \ref{fig_few_shot}. By gradually reducing the students' response logs, we simulate the model performance in the scenarios of limited knowledge coverage. The results show that as the coverage rate decreases, the performance decline of our model is noticeably smaller than that of other models. This phenomenon highlights the strength of our model in addressing the knowledge coverage problem. The concept-aware multidimensional vector representation captures richer semantic information, enhancing the representation capacity in sparse coverage scenarios. Additionally, the integration of the latent Q-matrix generated through attention-based knowledge aggregation with explicit knowledge effectively alleviates the sparsity issue of the Q-matrix. These components enable the model to extract valuable insights from limited data, enhancing its robustness with incomplete information.

\subsection{Rationality Analysis of Diagnosis Results}

\begin{figure}[t]
    \centering
	\includegraphics[width=.8\linewidth, height=.5\textheight,keepaspectratio=true]{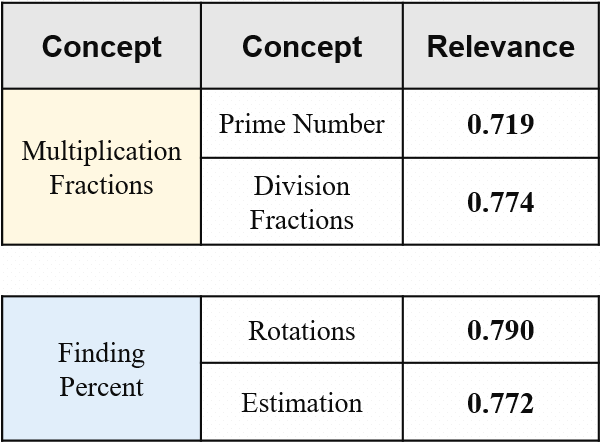}
    \caption{The results of student diagnosis report.}
    \label{fig_Rationality_Analysis}
\end{figure}

To analyze the validity of the CLEKI-CD diagnostic results, we randomly select a student from the ASSIST dataset for a case study. In Figure \ref{fig_Rationality_Analysis}, each row displays the chosen knowledge concept, associated exercise records, relevance (computed from the concept similarity matrix) to concepts with colored bottom frames and the mastery level diagnosed by the model. From Figure \ref{fig_Rationality_Analysis}, the student demonstrates strong mastery in "Prime Number" and "Division Fractions", evidenced by correct responses to most related exercises during training. Notably, the model also infers high mastery in "Multiplication Fractions" even though exercises on it are not included in the training, which can be attributed to the high relevance between "Multiplication Fractions" and "Prime Number" as well as "Division Fractions". The student's performance on test set exercises further supports this inference. Similarly, the model infers the low level of mastery in "Rounding," "Estimation," and "Finding Percent," which is consistent with the student's limited response logs and higher error rates. In the open test, students often encounter unseen concepts, leading to knowledge coverage issues. CLEKI-CD infers mastery of unpracticed concepts based on concept correlations, providing a more comprehensive cognitive diagnosis with strong foresight.

\subsection{Hyperparameter Analysis}

\begin{figure}[bt]
    \centering
    \begin{subfigure}[t]{0.49\linewidth}
        \centering
        \includegraphics[width=\linewidth]{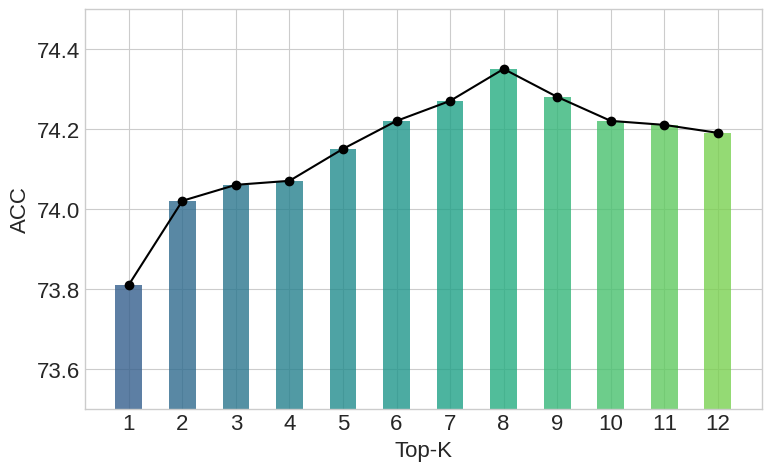}
        \caption{}
        \label{fig:top_k}
    \end{subfigure}
    \hfill
    \begin{subfigure}[t]{0.49\linewidth}
        \centering
        \includegraphics[width=\linewidth]{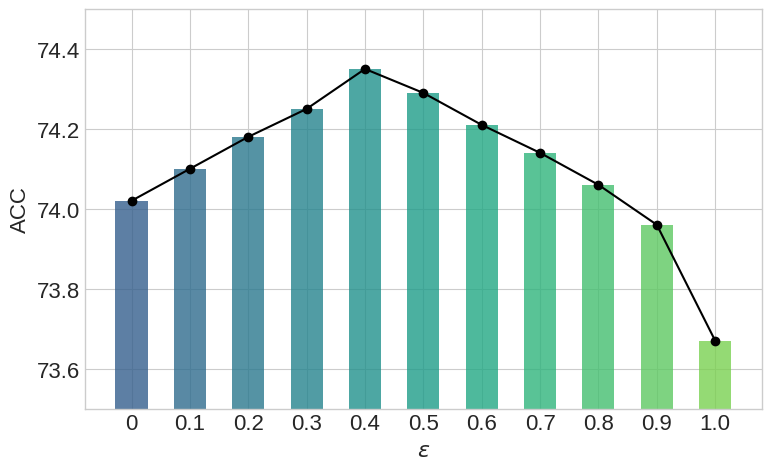}
        \caption{}
        \label{fig:epsilon}
    \end{subfigure}
    \caption{Hyperparameter analysis of (a) top-k and (b) $\epsilon$.}
    \label{fig:hyper_analysis}
\end{figure}

We demonstrate the impact of pivotal hyperparameters: top-k and the balance coefficient $\epsilon$ in the diagnostic layer using the Assist dataset. As illustrated in Figure \ref{fig:hyper_analysis} (a), the performance initially improves as more latent knowledge is added with increasing $k$, but declines once $k$ becomes too large due to introducing noise. The optimal performance is achieved with the top-8 latent concepts to supplement the sparse Q-matrix. The balance coefficient $\epsilon$ analyzed in Figure \ref{fig:hyper_analysis} (b) is used to balance the contributions of explicit and latent knowledge in the diagnosis process. When $\epsilon$ is set too close to 0 or 1, the model predominantly considers one type of knowledge, which limits its diagnostic comprehensiveness. This finding aligns with the results of the ablation study, which has shown that cognitive diagnosis using only one kind of knowledge reduces the performance.

\section{Conclusion}
In this work, we propose a novel cognitive diagnosis model termed CLEKI-CD, which aims to improve the accuracy and generalization of cognitive diagnosis. Unlike existing CDMs that represent students' mastery and exercise difficulty on a specific concept from a unidimensional perspective, we extend the representation into a multidimensional form that captures richer semantic information and establishes interactions within hierarchical relationships. Furthermore, an attention-based knowledge aggregation method is employed to leverage the inherent relationships among knowledge concepts to generate a latent Q-matrix that supplements the original sparse Q-matrix and alleviates the knowledge coverage problem. A combined cognitive diagnosis layer is then used to integrate both latent and explicit knowledge for a more comprehensive diagnosis. Extensive experiments on real-world datasets demonstrate significant improvements over state-of-the-art models, with case studies highlighting the interpretability and diagnostic capabilities. Future work will explore the incorporation of multi-modal features to further expand the applicability of the model.

\bibliographystyle{named}
\bibliography{ijcai25}

\begin{thebibliography}{}

\bibitem[\protect\citeauthoryear{Ackerman}{2014}]{ackerman2014multidimensional}
Terry~A Ackerman.
\newblock Multidimensional item response theory models.
\newblock {\em Wiley StatsRef: Statistics Reference Online}, 2014.

\bibitem[\protect\citeauthoryear{Bu \bgroup \em et al.\egroup }{2022}]{bu2022cognitive}
Chenyang Bu, Fei Liu, Zhiyong Cao, Lei Li, Yuhong Zhang, Xuegang Hu, and Wenjian Luo.
\newblock Cognitive diagnostic model made more practical by genetic algorithm.
\newblock {\em IEEE Transactions on Emerging Topics in Computational Intelligence}, 7(2):447--461, 2022.

\bibitem[\protect\citeauthoryear{Chang \bgroup \em et al.\egroup }{2015}]{chang2015modeling}
Haw-Shiuan Chang, Hwai-Jung Hsu, Kuan-Ta Chen, et~al.
\newblock Modeling exercise relationships in e-learning: A unified approach.
\newblock In {\em EDM}, pages 532--535, 2015.

\bibitem[\protect\citeauthoryear{Cheng \bgroup \em et al.\egroup }{2019}]{cheng2019dirt}
Song Cheng, Qi~Liu, Enhong Chen, Zai Huang, Zhenya Huang, Yiying Chen, Haiping Ma, and Guoping Hu.
\newblock Dirt: Deep learning enhanced item response theory for cognitive diagnosis.
\newblock In {\em Proceedings of the 28th ACM international conference on information and knowledge management}, 2019.

\bibitem[\protect\citeauthoryear{De~La~Torre}{2009}]{de2009dina}
Jimmy De~La~Torre.
\newblock Dina model and parameter estimation: A didactic.
\newblock {\em Journal of educational and behavioral statistics}, 34(1):115--130, 2009.

\bibitem[\protect\citeauthoryear{Gao \bgroup \em et al.\egroup }{2021}]{gao2021rcd}
Weibo Gao, Qi~Liu, Zhenya Huang, Yu~Yin, Haoyang Bi, Mu-Chun Wang, Jianhui Ma, Shijin Wang, and Yu~Su.
\newblock Rcd: Relation map driven cognitive diagnosis for intelligent education systems.
\newblock In {\em Proceedings of the 44th international ACM SIGIR conference on research and development in information retrieval}, pages 501--510, 2021.

\bibitem[\protect\citeauthoryear{Gao \bgroup \em et al.\egroup }{2023}]{gao2023leveraging}
Weibo Gao, Hao Wang, Qi~Liu, Fei Wang, Xin Lin, Linan Yue, Zheng Zhang, Rui Lv, and Shijin Wang.
\newblock Leveraging transferable knowledge concept graph embedding for cold-start cognitive diagnosis.
\newblock In {\em Proceedings of the 46th international ACM SIGIR conference on research and development in information retrieval}, pages 983--992, 2023.

\bibitem[\protect\citeauthoryear{Huang \bgroup \em et al.\egroup }{2020}]{huang2020learning}
Zhenya Huang, Qi~Liu, Yuying Chen, Le~Wu, Keli Xiao, Enhong Chen, Haiping Ma, and Guoping Hu.
\newblock Learning or forgetting? a dynamic approach for tracking the knowledge proficiency of students.
\newblock {\em ACM Transactions on Information Systems (TOIS)}, 38(2):1--33, 2020.

\bibitem[\protect\citeauthoryear{Huang \bgroup \em et al.\egroup }{2021}]{huang2021group}
Jie Huang, Qi~Liu, Fei Wang, Zhenya Huang, Songtao Fang, Runze Wu, Enhong Chen, Yu~Su, and Shijin Wang.
\newblock Group-level cognitive diagnosis: A multi-task learning perspective.
\newblock In {\em 2021 IEEE International Conference on Data Mining (ICDM)}, pages 210--219. IEEE, 2021.

\bibitem[\protect\citeauthoryear{Leighton \bgroup \em et al.\egroup }{2004}]{leighton2004attribute}
Jacqueline~P Leighton, Mark~J Gierl, and Stephen~M Hunka.
\newblock The attribute hierarchy method for cognitive assessment: A variation on tatsuoka's rule-space approach.
\newblock {\em Journal of educational measurement}, 41(3):205--237, 2004.

\bibitem[\protect\citeauthoryear{Li \bgroup \em et al.\egroup }{2022a}]{li2022hiercdf}
Jiatong Li, Fei Wang, Qi~Liu, Mengxiao Zhu, Wei Huang, Zhenya Huang, Enhong Chen, Yu~Su, and Shijin Wang.
\newblock Hiercdf: A bayesian network-based hierarchical cognitive diagnosis framework.
\newblock In {\em Proceedings of the 28th ACM SIGKDD conference on knowledge discovery and data mining}, pages 904--913, 2022.

\bibitem[\protect\citeauthoryear{Li \bgroup \em et al.\egroup }{2022b}]{li2022cognitive}
Sheng Li, Quanlong Guan, Liangda Fang, Fang Xiao, Zhenyu He, Yizhou He, and Weiqi Luo.
\newblock Cognitive diagnosis focusing on knowledge concepts.
\newblock In {\em Proceedings of the 31st ACM International Conference on Information \& Knowledge Management}, pages 3272--3281, 2022.

\bibitem[\protect\citeauthoryear{Liu \bgroup \em et al.\egroup }{2018}]{liu2018fuzzy}
Qi~Liu, Runze Wu, Enhong Chen, Guandong Xu, Yu~Su, Zhigang Chen, and Guoping Hu.
\newblock Fuzzy cognitive diagnosis for modelling examinee performance.
\newblock {\em ACM Transactions on Intelligent Systems and Technology (TIST)}, 9(4):1--26, 2018.

\bibitem[\protect\citeauthoryear{Liu \bgroup \em et al.\egroup }{2023}]{liu2023qccdm}
Shuo Liu, Hong Qian, Mingjia Li, and Aimin Zhou.
\newblock Qccdm: A q-augmented causal cognitive diagnosis model for student learning.
\newblock In {\em ECAI 2023}, pages 1536--1543. IOS Press, 2023.

\bibitem[\protect\citeauthoryear{Liu \bgroup \em et al.\egroup }{2024}]{liu2024inductive}
Shuo Liu, Junhao Shen, Hong Qian, and Aimin Zhou.
\newblock Inductive cognitive diagnosis for fast student learning in web-based intelligent education systems.
\newblock In {\em Proceedings of the ACM on Web Conference 2024}, pages 4260--4271, 2024.

\bibitem[\protect\citeauthoryear{Liu}{2021}]{liu2021towards}
Qi~Liu.
\newblock Towards a new generation of cognitive diagnosis.
\newblock In {\em The International Joint Conference on Artificial Intelligence}, 2021.

\bibitem[\protect\citeauthoryear{Lu \bgroup \em et al.\egroup }{2024}]{lu2024design}
Yu~Lu, Deliang Wang, Penghe Chen, and Zhi Zhang.
\newblock Design and evaluation of trustworthy knowledge tracing model for intelligent tutoring system.
\newblock {\em IEEE Transactions on Learning Technologies}, 2024.

\bibitem[\protect\citeauthoryear{Ma \bgroup \em et al.\egroup }{2022}]{ma2022knowledge}
Haiping Ma, Manwei Li, Le~Wu, Haifeng Zhang, Yunbo Cao, Xingyi Zhang, and Xuemin Zhao.
\newblock Knowledge-sensed cognitive diagnosis for intelligent education platforms.
\newblock In {\em Proceedings of the 31st ACM international conference on information \& knowledge management}, pages 1451--1460, 2022.

\bibitem[\protect\citeauthoryear{Ma \bgroup \em et al.\egroup }{2024}]{ma2024dgcd}
Haiping Ma, Siyu Song, Chuan Qin, Xiaoshan Yu, Limiao Zhang, Xingyi Zhang, and Hengshu Zhu.
\newblock Dgcd: an adaptive denoising gnn for group-level cognitive diagnosis.
\newblock In {\em The International Joint Conference on Artificial Intelligence}, 2024.

\bibitem[\protect\citeauthoryear{Myerson \bgroup \em et al.\egroup }{2001}]{myerson2001area}
Joel Myerson, Leonard Green, and Missaka Warusawitharana.
\newblock Area under the curve as a measure of discounting.
\newblock {\em Journal of the experimental analysis of behavior}, 76(2):235--243, 2001.

\bibitem[\protect\citeauthoryear{Qian \bgroup \em et al.\egroup }{2024}]{qian2024orcdf}
Hong Qian, Shuo Liu, Mingjia Li, Bingdong Li, Zhi Liu, and Aimin Zhou.
\newblock Orcdf: An oversmoothing-resistant cognitive diagnosis framework for student learning in online education systems.
\newblock In {\em Proceedings of the 30th ACM SIGKDD Conference on Knowledge Discovery and Data Mining}, pages 2455--2466, 2024.

\bibitem[\protect\citeauthoryear{Schlichtkrull \bgroup \em et al.\egroup }{2017}]{schlichtkrull2017modeling}
Michael Schlichtkrull, Thomas~N Kipf, Peter Bloem, Rianne van~den Berg, Ivan Titov, and Max Welling.
\newblock Modeling relational data with graph convolutional networks.
\newblock {\em arXiv preprint arXiv:1703.06103}, 2017.

\bibitem[\protect\citeauthoryear{Shao \bgroup \em et al.\egroup }{2024}]{shao2024improving}
Pengyang Shao, Chen Gao, Lei Chen, Yonghui Yang, Kun Zhang, and Meng Wang.
\newblock Improving cognitive diagnosis models with adaptive relational graph neural networks.
\newblock {\em arXiv preprint arXiv:2403.05559}, 2024.

\bibitem[\protect\citeauthoryear{Song \bgroup \em et al.\egroup }{2023}]{song2023deep}
Lingyun Song, Mengting He, Xuequn Shang, Chen Yang, Jun Liu, Mengzhen Yu, and Yu~Lu.
\newblock A deep cross-modal neural cognitive diagnosis framework for modeling student performance.
\newblock {\em Expert Systems with Applications}, 230:120675, 2023.

\bibitem[\protect\citeauthoryear{Tong \bgroup \em et al.\egroup }{2021}]{tong2021item}
Shiwei Tong, Qi~Liu, Runlong Yu, Wei Huang, Zhenya Huang, Zachary~A Pardos, and Weijie Jiang.
\newblock Item response ranking for cognitive diagnosis.
\newblock In {\em The International Joint Conference on Artificial Intelligence}, 2021.

\bibitem[\protect\citeauthoryear{Veli{\v{c}}kovi{\'c} \bgroup \em et al.\egroup }{2018}]{velivckovic2018graph}
Petar Veli{\v{c}}kovi{\'c}, Guillem Cucurull, Arantxa Casanova, Adriana Romero, Pietro Li{\`o}, and Yoshua Bengio.
\newblock Graph attention networks.
\newblock In {\em International Conference on Learning Representations}, 2018.

\bibitem[\protect\citeauthoryear{Wang \bgroup \em et al.\egroup }{2020}]{wang2020neural}
Fei Wang, Qi~Liu, Enhong Chen, Zhenya Huang, Yuying Chen, Yu~Yin, Zai Huang, and Shijin Wang.
\newblock Neural cognitive diagnosis for intelligent education systems.
\newblock In {\em Proceedings of the AAAI conference on artificial intelligence}, volume~34, pages 6153--6161, 2020.

\bibitem[\protect\citeauthoryear{Wang \bgroup \em et al.\egroup }{2022}]{wang2022neuralcd}
Fei Wang, Qi~Liu, Enhong Chen, Zhenya Huang, Yu~Yin, Shijin Wang, and Yu~Su.
\newblock Neuralcd: a general framework for cognitive diagnosis.
\newblock {\em IEEE Transactions on Knowledge and Data Engineering}, 35(8):8312--8327, 2022.

\bibitem[\protect\citeauthoryear{Wang \bgroup \em et al.\egroup }{2023}]{wang2023self}
Shanshan Wang, Zhen Zeng, Xun Yang, and Xingyi Zhang.
\newblock Self-supervised graph learning for long-tailed cognitive diagnosis.
\newblock In {\em Proceedings of the AAAI conference on artificial intelligence}, volume~37, pages 110--118, 2023.

\bibitem[\protect\citeauthoryear{Wang \bgroup \em et al.\egroup }{2024}]{wang2024survey}
Fei Wang, Weibo Gao, Qi~Liu, Jiatong Li, Guanhao Zhao, Zheng Zhang, Zhenya Huang, Mengxiao Zhu, Shijin Wang, Wei Tong, et~al.
\newblock A survey of models for cognitive diagnosis: New developments and future directions.
\newblock {\em arXiv preprint arXiv:2407.05458}, 2024.

\bibitem[\protect\citeauthoryear{Willmott and Matsuura}{2005}]{willmott2005advantages}
Cort~J Willmott and Kenji Matsuura.
\newblock Advantages of the mean absolute error (mae) over the root mean square error (rmse) in assessing average model performance.
\newblock {\em Climate research}, 30(1):79--82, 2005.

\bibitem[\protect\citeauthoryear{Yang \bgroup \em et al.\egroup }{2016}]{yang2016hierarchical}
Zichao Yang, Diyi Yang, Chris Dyer, Xiaodong He, Alex Smola, and Eduard Hovy.
\newblock Hierarchical attention networks for document classification.
\newblock In {\em Proceedings of the 2016 conference of the North American chapter of the association for computational linguistics: human language technologies}, pages 1480--1489, 2016.

\bibitem[\protect\citeauthoryear{Yen and Fitzpatrick}{2006}]{yen2006item}
Wendy~M Yen and Anne~R Fitzpatrick.
\newblock Item response theory.
\newblock {\em Educational measurement}, 4:111--153, 2006.

\end{thebibliography}

\end{document}